\documentclass[10pt,twocolumn,letterpaper]{article}

\usepackage{cvpr}
\usepackage{times}
\usepackage{epsfig}
\usepackage{graphicx}
\usepackage{amsmath}
\usepackage{amssymb}

\usepackage{hhline}
\usepackage{multicol}
\usepackage{arydshln}
\usepackage{overpic}
\usepackage{multirow}

\usepackage[breaklinks=true,bookmarks=false]{hyperref}

\cvprfinalcopy 

\def\cvprPaperID{****} 

\begin{document}

\title{Efficient Progressive Neural Architecture Search\thanks{To be presented at the \emph{British Machine Vision Conference} (BMVC 2018)}}


\author{
	\begin{tabular*}{0.9\linewidth}{@{\extracolsep{\fill}}ccc}
		~~Juan-Manuel P\'erez-R\'ua~~ & ~~~Moez Baccouche~~~ & St\'ephane Pateux\\		
		{\tt\small juanmanuel.perezrua@orange.com} &
		{\tt\small moez.baccouche@orange.com} & 
		{\tt\small stephane.pateux@orange.com}\\
		\multicolumn{3}{c}{\newline}\\		
		\multicolumn{3}{c}{Orange Labs}\\
		\multicolumn{3}{c}{4 rue Clos Courtel, 35512}\\
		\multicolumn{3}{c}{Cesson-S\'evign\'e, FR}
	\end{tabular*}
}

\maketitle

\begin{abstract}
This paper addresses the difficult problem of finding an optimal neural architecture design for a given image classification task. We propose a method that aggregates two main results of the previous state-of-the-art in neural architecture search. These are, appealing to the strong sampling efficiency of a search scheme based on sequential model-based optimization (SMBO)~\cite{liu2017progressive}, and increasing training efficiency by sharing weights among sampled architectures~\cite{pham2018efficient}. Sequential search has previously demonstrated its capabilities to find state-of-the-art neural architectures for image classification. However, its computational cost remains high, even unreachable under modest computational settings. Affording SMBO with weight-sharing alleviates this problem. On the other hand, progressive search with SMBO is inherently greedy, as it leverages a learned surrogate function to predict the validation error of neural architectures. This prediction is directly used to rank the sampled neural architectures. We propose to attenuate the greediness of the original SMBO method by relaxing the role of the surrogate function so it predicts architecture sampling probability instead. We demonstrate with experiments on the CIFAR-10 dataset that our method, denominated \emph{Efficient progressive neural architecture search} (EPNAS), leads to increased search efficiency, while retaining competitiveness of found architectures. 

\end{abstract}

\section{Introduction}
\label{sec:intro}

Since the popularization of convolutional neural networks (CNN) for image classification by Krizhevsky \etal~\cite{krizhevsky2012imagenet} in 2012, many subsequent works have proposed new hand-designed architectures leading to steady performance improvements over the years (see, for example~\cite{he2015delving, he2016deep, simonyan2015very, szegedy2015going, huang2017densely}, among others). The process leading to all these discoveries is almost always arduous, requiring careful experimentation and the intuition of an expert. Some of these findings seem to point to the fact that intricate neural network designs can offer large gains in performance~\cite{he2016deep, huang2017densely, szegedy2015going}. However, it is not always obvious what to change in a known design to push performance even further. Evidently, there is a need for efficient yet effective ways to discover new architectures of CNN automatically. This is the problem that we tackle in this paper. Here, we focus on finding optimal CNN architectures for image classification, while keeping in mind that, very often, advances in neural networks for image classification transfer to a large variety of other learning problems.

Current advances in neural architecture optimization can be categorized in roughly three groups: genetic algorithms (GA), reinforcement learning (RL), and surrogate-based optimization (SO). GA-based approaches~\cite{real2017large, stanley2002evolving,xie2017genetic} consist on iteratively mutating, training and evaluating promising architectures, while only top performers on a validation set are selected for further mutation. In RL methods~\cite{zhong2018practical, zoph2017neural, zoph2017learning}, a controller agent generates the description of a neural model, which is then trained and evaluated on a validation set. Validation error is then fed-back to the agent as a reward function, improving the controller and generating better models in future iterations. On the other hand, by SO we refer to methods that rely on learning a sort of surrogate function expressing a relationship between sampled models and validation error or similar. Typical examples are Bayesian optimization~\cite{snoek2012practical, snoek2015scalable} and sequential model-based optimization (SMBO)~\cite{liu2017progressive}. 

Recently, progressive neural architecture search (PNAS)~\cite{liu2017progressive}, one of the surrogate-based search methods, achieved state-of-the-art results on the CIFAR-10 dataset~\cite{krizhevsky2009learning}. Their algorithm performs a progressive scan of the neural architecture search space (which is constrained by design according to findings of previous state-of-the-art). The top K best performing architectures are chosen at each step of the algorithm and validation errors are collected by training the selected architectures for several epochs. These errors are then used to train a surrogate function that predicts validation error of subsequent architectures. The surrogate function allows efficient exploration of the search space by decreasing the amount of architectures that are actually trained. Computational cost is nonetheless high, requiring 100 GPUs working for 2 days to achieve their best results. However, PNAS~\cite{liu2017progressive} is considerably more efficient than previous methods, which rely on up to 800 GPUs working for a month (\ie, NAS~\cite{zoph2017neural}) to achieve their best results. 

A recent work based on RL, ENAS~\cite{pham2018efficient}, proposed to leverage weight-sharing among sampled architectures to increase search efficiency. The motivation behind this idea is the observation that the main bottleneck during neural architecture search is the training of sampled models to completion. Weight-sharing in this context removes the need to train sampled architectures from scratch. This idea improves time efficiency by a factor of $1000$ with respect to NAS~\cite{zoph2017neural}. We propose in this paper \textbf{to investigate the effect on speed and performance of weight-sharing in progressive SMBO-based optimization} for neural architecture search. The purpose of this is to benefit from the simplicity and impressive performance of PNAS~\cite{liu2017progressive}, while improving on speed thanks to the efficiency of ENAS~\cite{pham2018efficient}. Leveraging the gain on efficiency, we also propose \textbf{to relax the sampling strategy of PNAS by performing probabilistic sampling of new architectures} based on the prediction of the surrogate function. We shall call this new approach \textbf{efficient progressive neural architecture search} (EPNAS).

The remainder of this paper is organized as follows. In Section~\ref{sec:related}, we give an overview of the related work and state-of-the-art in neural network architecture search. In Section~\ref{sec:ss} we introduce the search space our method acts on, and in Section~\ref{sec:core} we explain EPNAS, our algorithm for neural architecture optimization. Later on, in Section~\ref{sec:experiments}, we present experimental results and analysis of the properties of our method. Finally, in Section~\ref{sec:conclusions}, we give concluding remarks and discuss future work.

\section{Related work}
\label{sec:related}

Optimization of neural network hyperparameters is a problem that has been tackled since the early nineties~\cite{angeline1994evolutionary, branke1995evolutionary}. Researchers quickly realized that, due to the extremely high computational demands of the problem, neural topology optimization would need to wait for the advent of higher-order parallel computing hardware and software~\cite{branke1995evolutionary}. Traditionally, up to this point, most of the attempts to tackle the problem were based on \textbf{randomized search methods}. A few years later, in 2002, Stanley and Milkkulainen~\cite{stanley2002evolving} proposed an evolutionary random search approach that solves jointly for optimal weights and network topology. In their work, topology mutation follows structured constraints and it is performed progressively.

Fast forward to recent years, evolutionary and genetic algorithms are still relevant. However, most modern methods focus on evolving the neural topology~\cite{miikkulainen2017evolving, real2018regularized, real2017large, xie2017genetic}, leaving the optimization of neural weights to gradient-based approaches. These methods, however, are reserved for large hardware set-ups with hundreds of GPUs running for weeks.

\begin{figure*}[tb]
	\centering
	\begin{center}
		\begin{tabular}{c}
			\includegraphics[width=0.85\linewidth]{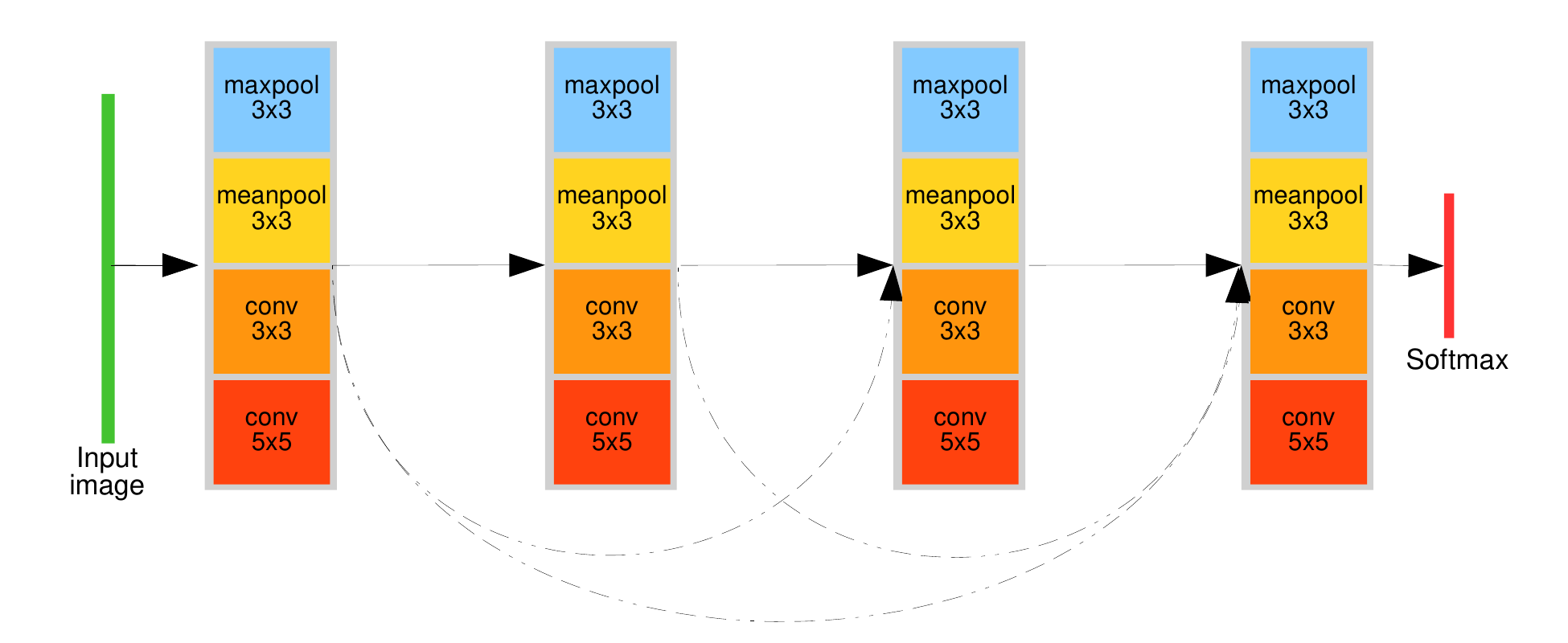} \\			
		\end{tabular}
	\end{center}
	\caption{\textbf{Structure of the macro search space used in this paper}. In this example, for the sake of simplicity, the number of layers is~$L=4$, and the number of possible operations per-layer is 4. The dotted lines represent the skip connections that are also part of the search space. 
	}
	\label{fig:search_space_macro}
\end{figure*}

Another group of practical approaches for automatically selecting a neural topology relies on \textbf{Bayesian optimization}~\cite{snoek2012practical, snoek2015scalable}. In this type of methods, the validation error of neural models is modeled as a Gaussian process, leading to optimal selection of candidate configurations. However, these methods cannot handle variable-size and variable-connectivity models.

Very recently, \textbf{reinforcement learning} appeared as an alternative to previous search methods. The technique usually consists of a controller recurrent neural network that samples new architectures at each iteration. This controller is trained with policy gradient~\cite{zoph2017neural} or Q-learning~\cite{zhong2018practical}, by feeding it with the validation errors collected by training the sampled networks to completion. A large scale benchmarking work by Real \etal, \cite{real2018regularized}, demonstrates that evolutionary approaches either match or surpass performance of RL methods while also reaching that outcome faster. In order to increase efficiency, subsequent works have focused on constraining the search space to modular elements composed of convolutional operations with demonstrated value for image recognition~\cite{zoph2017learning}. Similarly to the method from \cite{real2017large}, which allows weight inheritance during exploration of the search space, ENAS~\cite{pham2018efficient} proposes an interesting idea: sharing weights among sampled architectures. The weight-sharing strategy in RL-based approaches offers competitive results while improving exploration speed by a large margin.

In parallel to RL-based approaches, a different type of methods appeared recently. \textbf{Surrogate aided exploration} has also demonstrated accuracy and efficiency for network hyperparameter optimization~\cite{liu2017progressive, negrinho2017deeparchitect}. Sequential model-based optimization (SMBO) lies at the core of such methods. SMBO~\label{hutter2011sequential} originated as a technique for general algorithm parameter optimization. The method itself does not require gradients, but it requires the training of an intermediate function that scores parameter configurations. During exploration, this learned intermediate function or surrogate, is evaluated across a large set of parameter configuration candidates. The top $K$ performing configurations according to the surrogate are chosen to continue sequential exploration for another iteration. In the context of machine learning, this method has been used to automatically construct ensembles of classifiers by Lacoste \etal,~\cite{lacoste2014sequential}. On a different line, an interesting surrogate-based method, denominated SMASH~\cite{brock2017smash}, displays strong results. The surrogate function of SMASH is not designed to rank tested configurations, but to predict the weights of sampled architectures. Through random search, SMASH is able to find competitive architectures in a number of datasets. 

More recently, Liu~\etal,\cite{liu2017progressive} proposed to use SMBO-based progressive search (PNAS) within a constrained search space as in~\cite{zoph2017learning}, achieving state-of-the-art results in image classification on CIFAR-10, while also being significantly less computationally intensive than~\cite{zoph2017learning}. The architecture hyperparameters learned for CIFAR-10 were successfully transferred to ImageNet with simple modifications. However, the method is still considerably slower ($\sim 200$ times) than ENAS~\cite{pham2018efficient}. In this paper we are interested in assessing the inclusion of weight-sharing in SMBO-based approaches. Beyond that, we believe PNAS suffers from over-greediness as it merely chooses the top $K$ architectures at each step of the progression of the method. In this paper we are also interested in checking whether surrogate-based sampling could lead to better results with respect to the greedy choice of configuration candidates.

\section{Search space structure}
\label{sec:ss}

\begin{figure*}[tb]
	\centering
	\begin{center}
		\begin{tabular}{c}
			\includegraphics[width=0.80\linewidth]{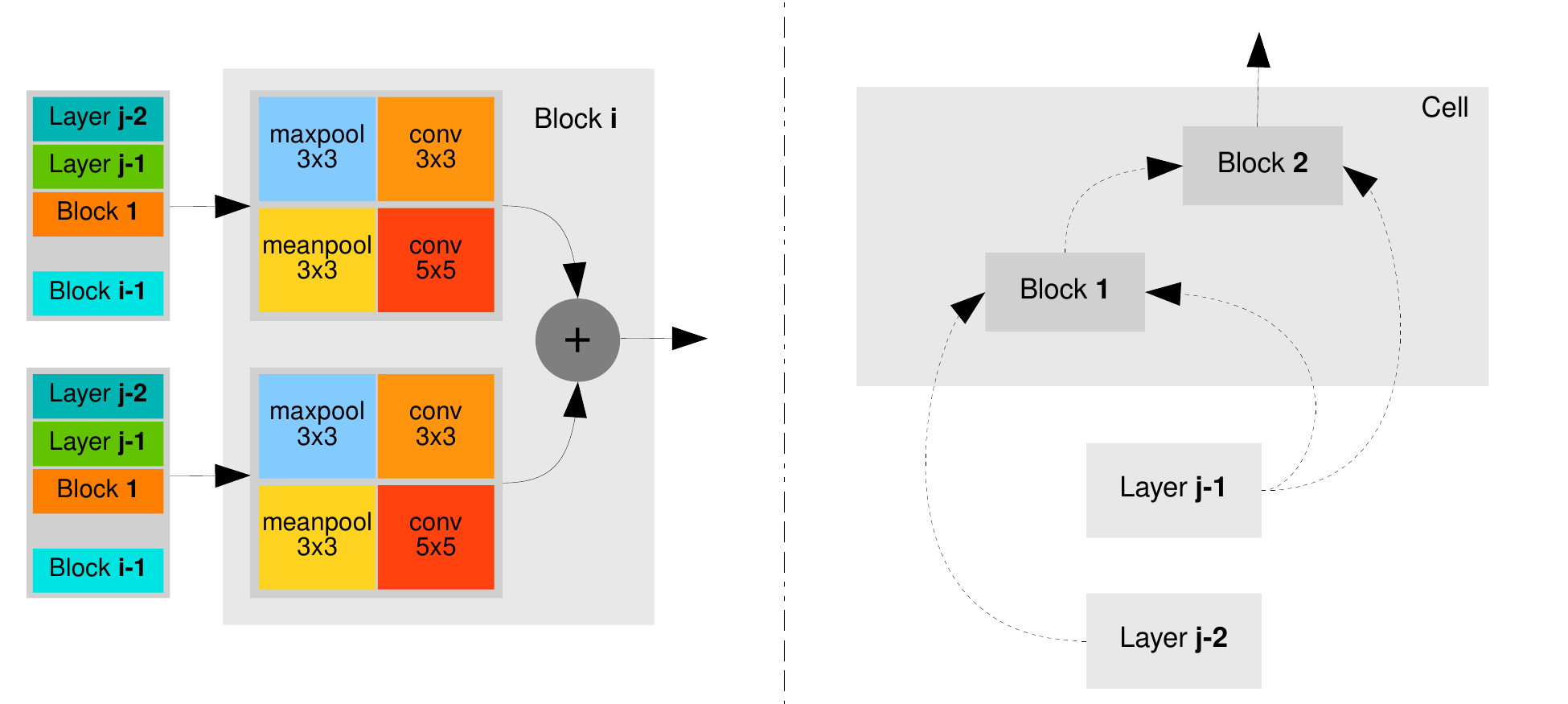} \\	
			\includegraphics[width=0.70\linewidth]{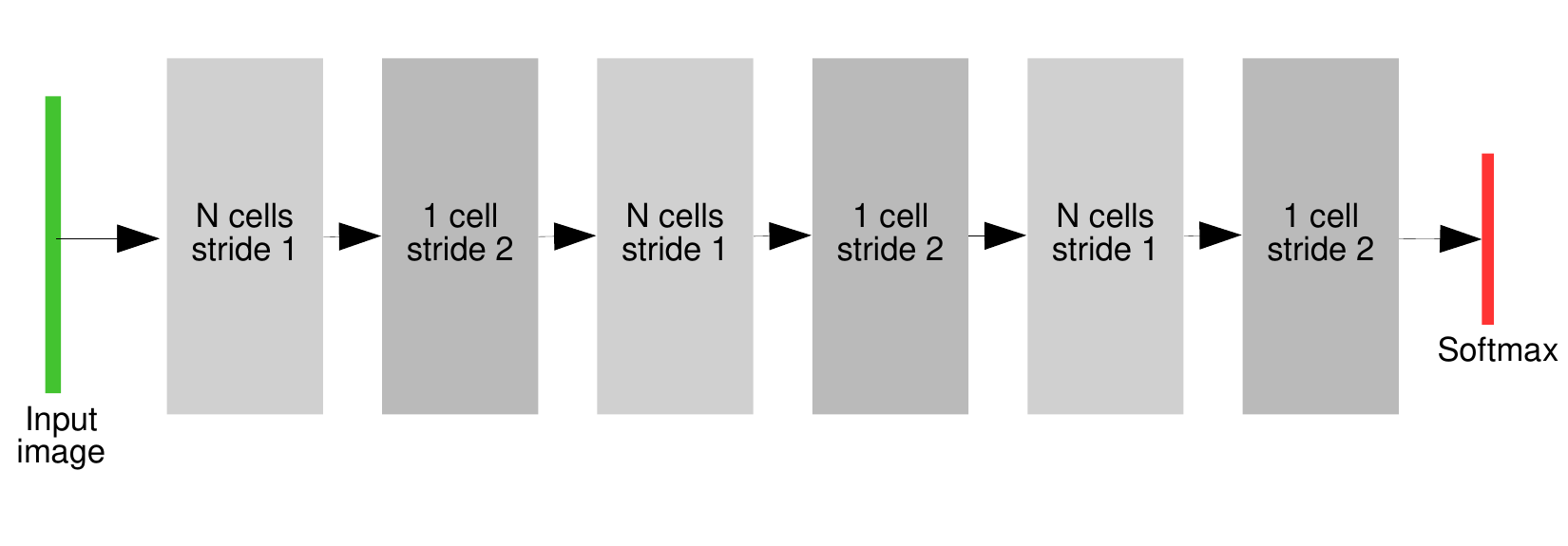}		
		\end{tabular}
	\end{center}
	\caption{\textbf{Structure of the micro search spaces used in this paper}. Top row: Block structure on the left, cell structure on the right. Recall that a cell is formed by one or more blocks. In this example, the number of possible operations per block is 4. Observe that for any given block inside a cell, the possible inputs for the two operations of the block are the outputs of previous two layers and previous blocks of the same cell. Bottom row: a neural network in the \emph{micro} search space is formed by stacks of cells with the same structure.
	}
	\label{fig:search_space_micro}
\end{figure*}

In this section, we describe the search space used by our method. Following~\cite{pham2018efficient}, we limit the search space of our algorithm to one of two possible structures. First, we consider what they denominate \emph{macro} search space, spanning entire deep convolutional models  with skip connections. Later on, we work on a \emph{micro} search space, operating over modular convolutional cells.

\paragraph{Macro search space.} First we focus on regular convolutional neural networks with variable skip connections and convolutional operations, namely the \emph{macro} space (see Fig.~\ref{fig:search_space_macro}). For this search space we start from a fixed convolutional module to expand the input image from 3 color channels to the fixed number of channels $C$ that will take place along the network, with a maximum number $L$ of convolutional layers. At each layer one of six possible operations is chosen. We follow~\cite{pham2018efficient} regarding the considered operations: 

\begin{itemize}
	\item $3\times 3$ max pooling\vspace{-2mm}

	\item $3\times 3$ mean pooling\vspace{-2mm}

	\item $3\times 3$ convolution\vspace{-2mm}

	\item $5\times 5$ convolution\vspace{-2mm}

	\item $3\times 3$ depthwise convolution~\cite{chollet2016xception}\vspace{-2mm}

	\item $5\times 5$ depthwise convolution~\cite{chollet2016xception}\vspace{-2mm}

\end{itemize}

At the output of the CNN described by a realization of the \emph{macro} search space, a softmax layer processes the channel-wise global average of the last convolutional layer. This idea, introduced by~\cite{lin2013network}, precludes the use of large fully connected layers that might prevent stable training during the architecture search. Overall, we can describe a layer $l$ of a network in the \emph{macro} space with a tuple $\left(O_l, S_l\right)$. Here, $O_i \in \mathcal{O}$, where $\mathcal{O}$ is the set of possible operations. $S_l$ is a variable-length tuple describing what previous layers are used as input to layer $l$. This allows the search space to contain neural networks with variable amounts of skip connections. Observe then that $S_l \subseteq (H_1,\cdots, H_{l-1})$, where $H_l$ is the output of layer $l$, avoiding unwanted loops in the underlying computational graph. To summarize, a whole network of $L$ layers is fully described by a $L$-tuple $\left((O_1, S_1),\cdots,(O_l, S_l),\cdots,(O_{L}, S_{L})\right)$.

\paragraph{Micro search space.} For the \emph{micro} search space, we keep the fixed input convolution as in the \emph{macro} space and employ, again, global average pooling of the last convolutional layer as input to the final softmax layer. Fig.~\ref{fig:search_space_micro} shows a diagram that explains the \emph{micro} search space in detail. A neural network configuration stemming from a realization of the \emph{micro} search space is formed by stacks of convolutional modules denominated \textbf{cells}. According to the network topology that is shown in the bottom row of Fig.~\ref{fig:search_space_micro}, a cell is replicated several times. In such a way, a whole network configuration is described only by a cell structure. 

As it can be appreciated in the bottom part of Fig.~\ref{fig:search_space_micro}, to form a CNN from the micro space, two types of cell are stacked: \emph{normal}, and \emph{reduction}. The only difference between them is that the reduction cells use strides. The objective is to increase the receptive field of deeper layers in the network. Normal cells are stacked $N$ times in between reduction ones.

\begin{figure*}[t]
	\centering
	\begin{overpic}[width=0.8\textwidth]{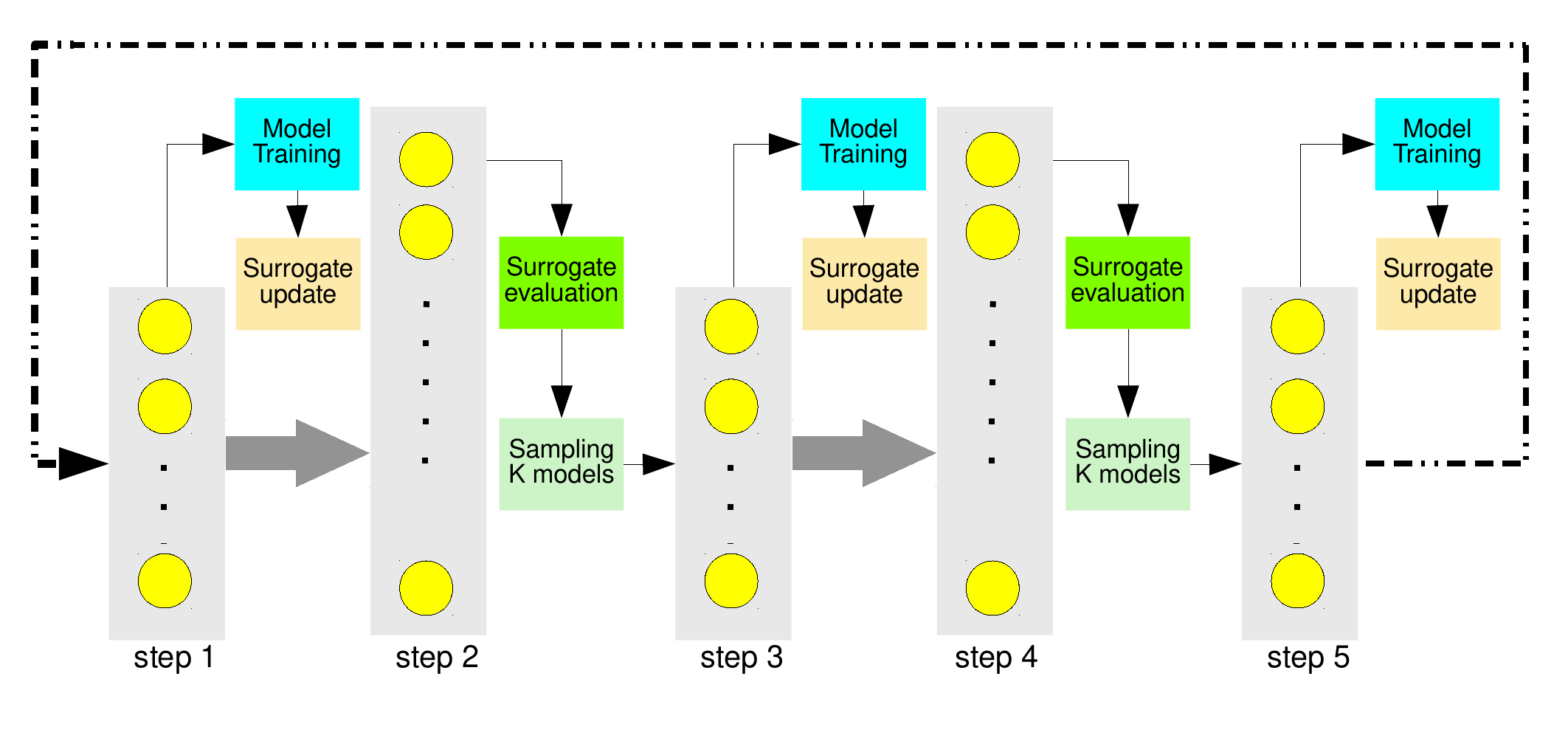}
	\end{overpic}
	\caption{\textbf{Illustration of the progressive model-based search procedure used in this paper.} In this example, a maximum number of three blocks (if considering the \emph{micro} space) or layers (\emph{macro} space) are used. \textbf{Step 1}: We list all the possible configurations for the simplest complexity level ($B=1$, or $L=1$). Each of these models are trained sequentially by sharing-weights among themselves, for a small number of epochs. When validation errors are collected, the surrogate function is trained for the first time. \textbf{Step 2}: The second level of complexity is unrolled, leading to a total number of possible configurations of $|\mathcal{B}_1|\times|\mathcal{B}_2|$. Instead of training all these configurations, the surrogate function is used to sample $K$ models. \textbf{Step 3}: $K$ configurations are trained, and the surrogate is updated. \textbf{Step 4}: The previously sampled $K$ configurations are combined with the new unroll of the architecture complexity. A subset of $\mathcal{B}_{1:3}$ is then evaluated by the surrogate and $K$ new configurations are sampled. \textbf{Step 5}: the surrogate is updated again. This whole procedure is repeated for a small number of iterations (dotted arrow) while cooling down the sampling temperature.}
	\label{fig:search}
\end{figure*}

Each one of these cells, as seen in the top row of Fig.~\ref{fig:search_space_micro} is formed by $B$ blocks. We shall denote the space of possible blocks with $\mathcal{B}$. As opposite to the depth-only structure of the \emph{macro} space, the block-based structure of the \emph{micro} space allows networks to go wider as needed. The structure of each block is defined by two convolutional operations, and their corresponding input connection. The possible operations for each block are the same ones than for the \emph{macro} search space plus the identity transformation\footnote{The identity operation allows blocks to directly combine features from previous layers and blocks in a similar way to ResNets\cite{he2016deep}.}. On the other hand, the possible input connections for a given block in a cell are the outputs of the previous two cells plus the outputs of all previous blocks inside the cell. Finally, the two convolutional features of a block are summed up, as in~\cite{liu2017progressive}. This means we can describe cell $c$ by a concatenation of $B$ $4$-tuples, one tuple with the form $(O_1, O_2, S_1, S_2)$ for each block $b$. For the \emph{micro} search space, $O_i \in \mathcal{O}$, and $S_i \in \{H_B^{c-2}, H_B^{c-1}, H_1^c,\cdots,H_{b-1}^c\}$. Here, $H^j_i$ is the output of block $i$ at cell $j$. When constructing the final cell output, we simply concatenate all the outputs of blocks that were not used as input of any other block within the cell.

The two described search spaces end up being extremely large. The cardinality of the solution space of \emph{macro} configuration is $\sim 2 \times 10^{29}$ when $L=12$. On the other hand, the solution set of the \emph{micro} space is of cardinality $\sim 5 \times 10^{14}$ when we allow up to $B=5$ blocks. To put those numbers in context, it can be noted that there are \emph{only} $\sim 2\times10^{11}$ galaxies containing $\sim 10^{24}$ stars in the observable universe.

\section{Efficient progressive neural architecture search}
\label{sec:core}

In this paper we adhere to the notion of PNAS~\cite{liu2017progressive} that highlights the difficulty of adequately exploring an exponential search space like ours in a direct manner. A more direct approach to neural architecture exploration as done by NAS~\cite{zoph2017learning} or ENAS~\cite{pham2018efficient} often requires sampling a large number of configurations. This is the main reason why we adopt a sequential exploration scheme.

\begin{table*}[tb]
	\centering 
	\caption{\textbf{Neural architecture search on the CIFAR10 dataset.} We present final validation errors for a number of methods, including our best architectures from the \emph{macro} and \emph{micro} search spaces. For the reported neural search methods, we provide with time duration of the search and the number of used GPUs.}
	
	\begin{tabular}{lcccc}
		
		\hline
		\multirow{ 2}{*}{Method} & \multirow{ 2}{*}{GPUs} & Time & Parameters & Error  \\
		& & (days) & (millions) &($\%$) \\
		\hhline{=====}
		ResNet~\cite{he2016deep} & - & - & 1.7 & 6.43\\		
		DenseNet~\cite{huang2017densely} & - & - & 25.6 & 3.46\\
		\hhline{=====}
		Super Nets \cite{VeniatD17} & - & - & - & 9.21\\
		ConvFabrics \cite{saxena2016convolutional} & - & - & 21.2 & 7.43 \\
		SMASH \cite{brock2017smash} & 1 & 1.5 & 16.0 & 4.03 \\
		QNAS \cite{baker2016designing} & 10 & 10 & 11.2 & 6.92 \\
		NAS \cite{zoph2017neural} & 800 & 28 & 37.4 & 3.65\\
		ENAS macro \cite{pham2018efficient} & 1 & 0.32 & 21.3 & 4.23 \\
		\hdashline[5pt/8pt]
		EPNAS macro (ours) & 1 & 1.2 & 5.9 & 5.14 \\
		EPNAS macro (ours + more channels) & 1 & 1.2 & 38.8 & 4.01 \\
		\hhline{=====}	
		NASNet micro \cite{zoph2017learning} & 450 & 4 & 3.3 & 3.41 \\
		ENAS micro \cite{pham2018efficient} & 1 & 0.45 & 4.6 & 3.54 \\
		PNAS micro \cite{liu2017progressive} & 100 & 1.5 & 3.2 & 3.63\\
		\hdashline[5pt/8pt]
		EPNAS micro (ours) & 1 & 1.8 & 1.6 & 5.69 \\
		
		EPNAS micro (ours + more channels) & 1 & 1.8 & 6.6 & 3.71 \\
		\hline
	\end{tabular}
	\label{tab:CIFAR10}
\end{table*}

\paragraph{Sequential search with shared weights.} Under such a paradigm, we explore the search space at increasing levels of model complexity, starting with simple ones first. This means that, in the case of the \emph{macro} search space we start the exploration by considering architectures with $L=1$, while for the \emph{micro} space we consider architectures with $B=1$. The sampled neural architectures are then trained for a small number of epochs\footnote{While \cite{pham2018efficient} only trains sampled architectures for a single epoch, we found our method to behave better by training for three epochs.}. This is possible because shared weights are refined further when new sampled architectures are trained\footnote{Two sampled networks share convolutional weights if the same convolutional operations at the same depth are used by the two of them.}. The accuracy scores derived from a validation dataset are then used to train a surrogate function that predicts scores of new configurations (details on the surrogate function as described later on). 

Subsequently, we take all the possible network configurations at the next level of complexity ($B=2$ for \emph{micro} and $L=2$ for \emph{macro}) and combine them with all the configurations of the previous step. At this point, the number of considered architectures is very high. Training all of them with modest hardware setups, even for very few epochs, would be too expensive. Instead, we sample $K$ of them. The probability of architecture $i$ with predicted accuracy $s_i$ to be selected is $p_i = s_i/\sum_{j}s_j$.

 Thus, our surrogate function predicts sampling probability of neural architectures. These $K$ sampled configurations are then trained again for the same number of epochs (three). The newly obtained validation accuracy scores are used to update the surrogate function. 

During following exploration steps, all the possible configurations that unfold at subsequent levels of complexity are combined with the currently sampled $K$ architectures. The \textit{sampling, training, and surrogate updating} loop continues until the maximum complexity level is reached. In Fig.~\ref{fig:search} we explain in more detail the search procedure of EPNAS.

Observe that, departing from PNAS~\cite{liu2017progressive}, we only train sampled architectures for a small number of epochs. The noisy estimation of sampled architecture accuracy is attenuated by implementing iterations on the sequential exploration. In our experiments, we repeat the whole process described before (the sequential exploration from simplest to highest level of network complexity) up to five times. More importantly, we adopt the graph-based interpretation of the search space proposed by~ENAS~\cite{pham2018efficient}, allowing us to employ their weight-sharing strategy. In this way, the more an operation at a layer or block is used by sampled architectures, the more its weights are updated. Effectively, search is biased towards architectures with previously trained modules as long as they have been refined more through gradient descent (presenting better classification accuracy scores). This is why in our implementation, probabilistic random sampling is preferred over simply taking the top K performing architectures. Random sampling allows our method to escape local minima. We implement a temperature-driven sampling procedure, so the search space is explored more randomly first, while fully trusting the surrogate function at the end. The final sampling probability of architecture $i$ is given by $p_i^{1/\tau}/\sum_{j}p_j^{1/\tau}$ with temperature $\tau$ decaying quickly to one as the number of iterations increases. In this sense, EPNAS lies in between a greedy progressive approach and purely random search.

\paragraph{Surrogate function.} The progressive search used in this paper, together with the type of architectures with variable length and connectivity we deal with, make of \emph{recurrent neural networks} an ideal candidate for implementing our surrogate function. This is a choice that is common in the literature~\cite{zoph2017learning, zhong2018practical, pham2018efficient, liu2017progressive}. In practice, we use a LSTM~\cite{hochreiter1997long} network. 

The actual input to the LSTM is computed with a small linear layer that acts as decoder for the symbol array describing the network configuration. Another linear layer with a sigmoid non-linearity processes the last hidden state of the LSTM. The output is a scalar in $[0,1]$ representing the accuracy score on the validation dataset of the input neural network described by the input string describing a neural configuration. The input dataloader for the LSTM is implemented with a hash-table so we can keep control of repeated network configurations. The surrogate function is trained with Adam~\cite{kingma2014adam} with a $L1$ loss.

\section{Experiments}
\label{sec:experiments}

\begin{figure*}[t]
	\centering
	\begin{overpic}[width=1.1\textwidth]{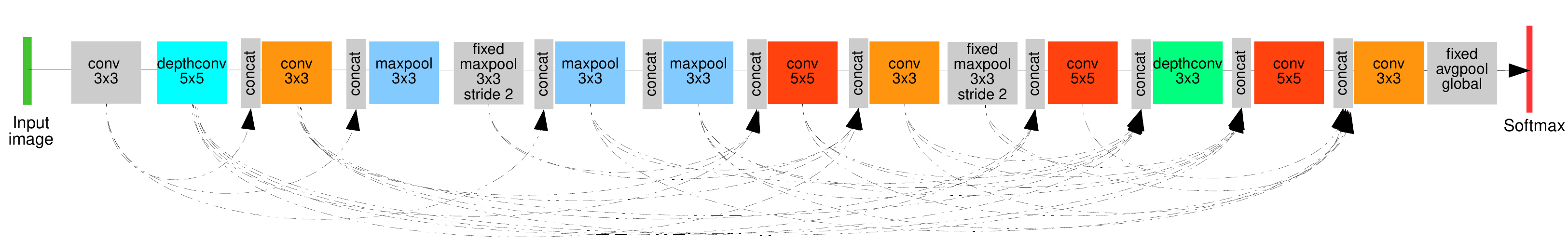}
	\end{overpic}
	\caption{\textbf{Best found CNN in the \emph{macro} search space.}}
	\label{fig:bestmacro}
\end{figure*}

\begin{figure*}[t]
	\centering
	\begin{overpic}[width=0.8\textwidth]{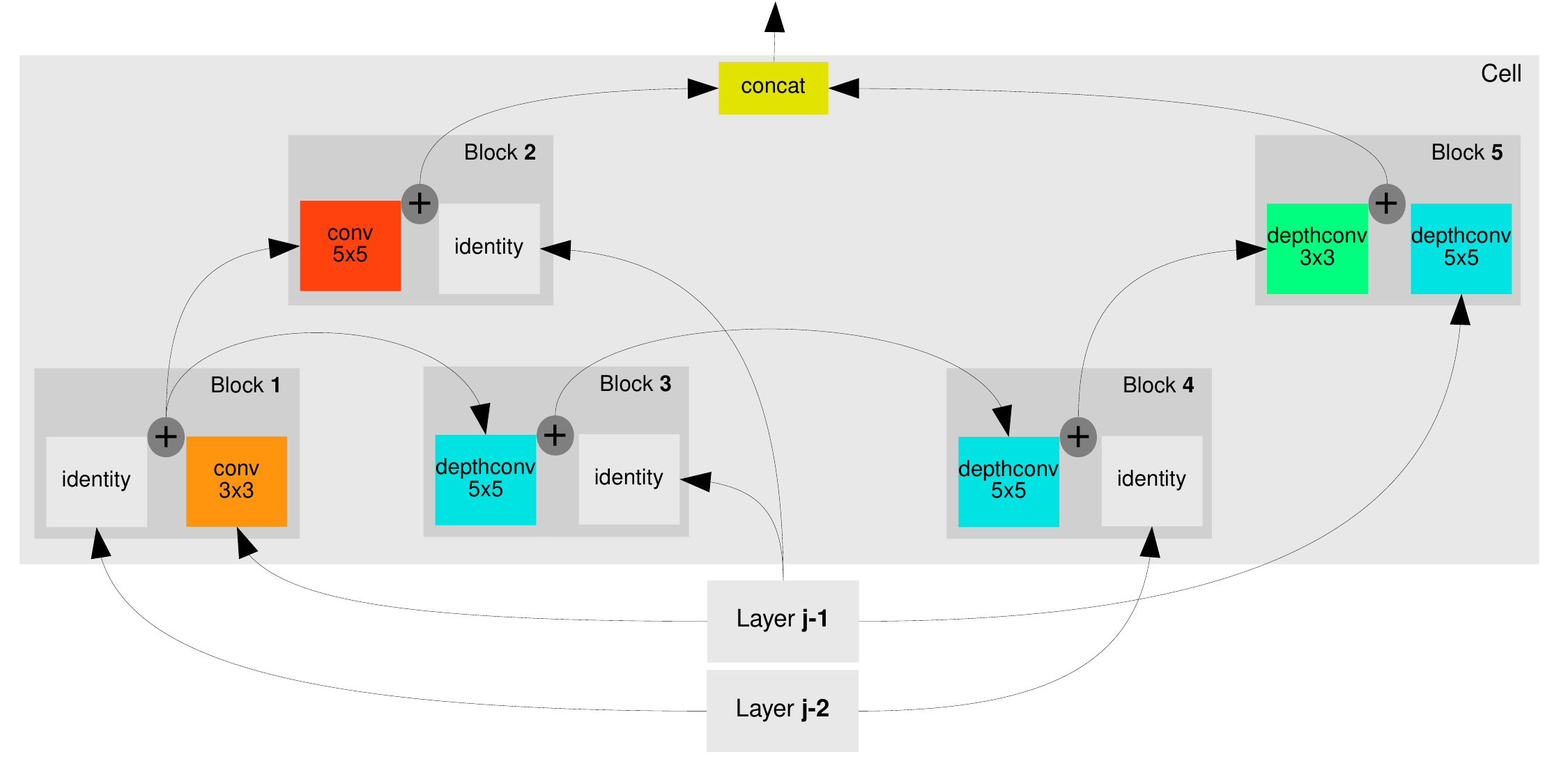}
	\end{overpic}
	\caption{\textbf{Best found cell in the \emph{micro} search space.}}
	\label{fig:bestmicro}
\end{figure*}

\paragraph{Experimental setup.} In this paper, neural architecture search for image classification is evaluated on CIFAR-10~\cite{krizhevsky2009learning}, as it is common practice~\cite{zoph2017learning, liu2017progressive}. This dataset contains 50.000 training images and testing 10.000 images. As it is standard for image classification, the images are normalized for training. The dataset is augmented by performing random-cropping and random mirroring. 

\paragraph{Implementation details.} Since we use shared weights, we need to guarantee that the number of input channels remain the same for operations at the same block or layer. We achieve this by including $1\times 1$ convolutions in between cells and layers, for the \emph{micro} and \emph{macro} search spaces respectively. This structure was originally proposed by \cite{pham2018efficient}. The weights of such intermediate convolutions are also shared among sampled architectures.

The surrogate function is a LSTM with 100 hidden units, processing an input feature of size 100. The weights of the linear layers are initialized randomly with a uniform distribution between $[-1,1]$.
During exploration, $K=25$ network configurations are sampled at each step. These configurations are trained sequentially for three epochs. The number of channels for the convolutional layers is 24 during exploration. For the \emph{micro} search space, the number of stacked normal cells is $N=2$, while the maximum number of blocks is $B=5$. During the final training of the found architectures, we let $N=3$ and the initial number of convolutional channels to 96 for the \emph{micro} space. On the other hand, for the \emph{macro} search space, we set the number of layers to $L=12$, and the initial number of channels to 200. We also evaluate the found configurations with a larger number of channels (512 for the \emph{macro} space, and 128 for the \emph{micro}). Models with fewer channels are faster but less accurate, while models with more channels turns into higher accuracy at the cost of being slower.
The convolutional weights are learned under the same optimizer configuration and learning rate schedule of~\cite{zoph2017learning}. Furthermore, we perform three to five iterations of our progressive search. In our experiments, three iterations always performed around $8\%$ better than a one-pass progressive search, even when the sampled models were trained for more epochs.

\paragraph{Discussion.} Our results are shown in Table~\ref{tab:CIFAR10} along with several baselines and other state-of-the-art algorithms. The first group in Table~\ref{tab:CIFAR10} presents results of recent hand-designed neural architectures. In particular, we show results by ResNet~\cite{he2016deep}, the work that introduced residual skip connections. Along with ResNets, impressive results for image classification by DenseNets~\cite{huang2017densely} are presented. The second group in Table~\ref{tab:CIFAR10} is conformed by other neural search methods, spanning several different kind of techniques. All of those methods aim at designing full networks, as in our \emph{macro} space. The third group is formed by more direct competitors of our method on the \emph{micro} search space. Indeed, PNAS~\cite{liu2017progressive} and NASNet~\cite{zoph2017learning} search neural architectures within a very similar search space to ours. 

\begin{figure}[tb]
	\centering
	\begin{center}
			\includegraphics[width=1\linewidth]{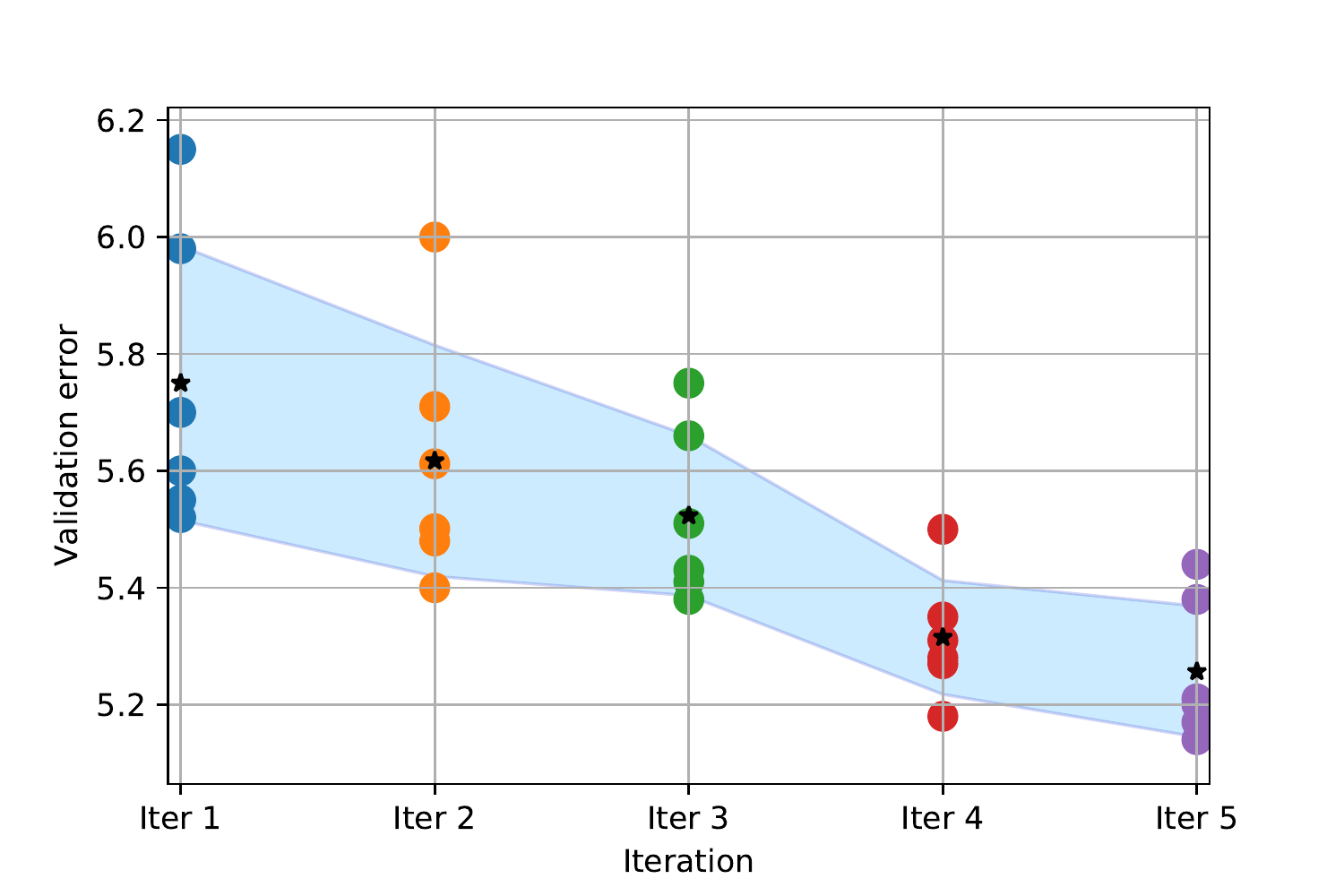} 
	\end{center}
	\caption{\textbf{Behaviour of EPNAS across multiple parallel runs and number of iterations on the \textit{macro} space.} Dots indicate independently run instances of EPNAS. Black star indicates the mean for runs with the same number of iterations and blue shadow the variance. 
	}
	\label{fig:beh}
\end{figure}

It must be observed that \textbf{our method delivers competitive results with respect to PNAS}, while being \textbf{close to a hundred times more efficient in terms of GPU/days}. The best found CNN from the \emph{macro} space is represented in Fig.~\ref{fig:bestmacro}, while the best found cell from the \emph{micro} space can be seen in Fig.~\ref{fig:bestmicro}. Moreover, even though NASNet~\cite{zoph2017learning} finds architectures that perform better on the validation set, our method is almost 900 times faster. Our results, not being too far from the absolute top performer, present convenient trade-off between number of parameters and performance, as it can be appreciated by our experiments with more channels.
It can also be pointed out that better results are expected with larger K. However, we noticed that the increase on exploration cost are large. With more than one available GPU, one could run several search threads in parallel and choose the best found architecture from all of the them. The behaviour of EPNAS among threads can be seen in Fig~\ref{fig:beh}. Observe that for few iterations the variance of the accuracy of found architectures is relatively high. This can be explained by the increased sampling temperature. As the number of iterations increases and the sampling temperature decreases, the variance is reduced. Observe that different architectures offer relatively similar validation accuracy. This is likely due to the shape of the constrained search space, which increases the likelihood of randomly sampled architectures to perform well. Overall, runs of EPNAS with more iterations seem to perform generally better.

\section{Conclusions}
\label{sec:conclusions}

In this paper we have introduced a new method for neural architecture search that works well on modest hardware configurations with a single or few GPUs only. Our method is based on a progressive paradigm by means of the \emph{Sequential Model-based Optimization} technique. To enable faster exploration, we adopt a recently-proposed heuristic from the state-of-the-art that consists of sharing weights among sampled architectures during exploration. This allows search methods to circumvent the need for complete training of sampled architectures, a common bottleneck in neural architecture search. The savings on computational expense allow us to explore modifications to the original sequential approach. We introduce search iterations and surrogate-based probabilistic sampling of network configurations as opposite to a one-shot greedy-selection of top architectures at each exploration step. We demonstrate on the challenging CIFAR-10 dataset that our method is able to find very competitive architectures with results that are not far from the state-of-the-art. With this paper, we also demonstrate the applicability of weight-sharing for neural architecture search methods based on progressive search. Future work includes the exploration of our efficient progressive neural architecture search (EPNAS) for a wider variety of learning problems. 

{\small
	\bibliographystyle{ieee}
	\bibliography{tex/biblio}
}

\end{document}